\documentclass[conference]{IEEEtran}
\usepackage{graphicx} 
\usepackage{soul}
\usepackage{booktabs}
\usepackage{xcolor}
\usepackage{amssymb}

\begin{document}

\pagestyle{plain}
\pagenumbering{arabic}

\title{Does Research-Paper Context Improve LLM-Generated Quadruped Navigation? An Exploratory Simulation Study}

\author{
\IEEEauthorblockN{Barnabas Juhasz}
\IEEEauthorblockA{
AlpineAI AG\\
Davos, Switzerland\\
}
\and
\IEEEauthorblockN{Rolf Pfister}
\IEEEauthorblockA{
Lab42 AI Research Institute\\
Davos, Switzerland\\
}
}

\maketitle 
\thispagestyle{plain}

\begin{abstract}
Large language models can write competent robot navigation code, but not only does the research on the topic advance, a model may also, due to the volume of findings, fail to recall relevant findings and apply them to the task at hand. In this work, we test whether providing particular research papers as context improves the LLM-written navigation code for the Go2 quadruped, with the LLM (Claude Opus 5) and locomotion policy frozen. Specifically, we compare the navigation controllers the LLM writes when provided with and without particular papers across three settings: behavior trees, navigators edited from a shared base, and controllers written from scratch. When a maze's shortest route is blocked mid-episode, 2 of 8 controllers written with the context of an incremental-replanning paper complete all four trips under Isaac Sim physics, against 7 of 8 without it (two-sided Fisher's exact test, p = 0.041). This difference appears only under physics: in an idealized kinematic simulator, the controllers written with and without the paper are indistinguishable. We hypothesize that an LLM provided research papers as context tends to overweight both their content and their goals - in this case, the efficiency of incremental replanning - negatively affecting the fulfillment of the main task and leading to less robust code. 
\end{abstract}

\section{Introduction}
Coding agents have become capable of writing competent robot navigation code \cite{liang2023codepolicieslanguagemodel, huang2023visuallanguagemapsrobot, vemprala2023chatgptroboticsdesignprinciples, fu2026capx, elmaaroufi2026rho}, but navigation still has open challenges \cite{Cadena_2016, xiao2022motionplanningcontrolmobile} and the research addressing them continues to advance, with scientific output growing by roughly $4\%$ per year \cite{bornmann2021growth}. Even a model trained on that research is unlikely to recall every relevant finding, or to think of applying it to the specific task in front of it. In particular, how reliably a model recalls a fact depends on how often it appears in its pre-training data \cite{kandpal2023longtail, mallen2023nottrust}, and its effective knowledge cutoff often precedes the reported one \cite{cheng2024dateddata}.
\\
\newline
Research literature could therefore complement the knowledge of a model in three ways. Firstly, \textit{novelty}: findings published after the training cutoff cannot be recalled at all. Secondly, \textit{coverage}: when asked plainly, a model tends to reproduce the few mechanisms most represented in its training data, while the literature contains a wide variety of alternatives. Thirdly, \textit{provenance}: a mechanism taken from a paper comes with its source and a falsifiable prediction, whereas a threshold stated by the model carries the same confidence whether or not it is grounded. We focus on research papers rather than documentation or code, as documentation tells the model how to call an interface \cite{zhou2023docprompting}, whereas a paper describes a mechanism and the objective it was designed to optimize. Although retrieved context is known to improve code generation, it can also go unused or degrade performance when it is long or only partially relevant \cite{wang2025coderagbench, shi2023distracted, du2025contextlength}. To the best of our knowledge, the net effect of research-paper context on robot controllers evaluated under physics has not yet been measured.
\\
\\
Our aim is therefore to test whether explicitly providing research findings to an LLM produces better navigation behavior on the Go2 robot \cite{unitreego2}, while keeping both the LLM and the locomotion policy fixed. Our research question is thus formulated as follows:

\noindent \textbf{RQ:} \textit{Would an LLM produce Go2 robot navigation code with a higher task-completion rate when given extra context from research literature?}

\noindent Specifically, in our first two settings, we measure task completion as the goal-reaching rate over $20$ generated maze worlds in a kinematic simulator. In our third and final setting, we measure whether a controller completes all four one-way trips within $180$\,s which we evaluate both in a kinematic simulator and under Isaac Sim physics.

\section{Methodology}
Throughout our different settings, the Go2 robot \cite{unitreego2} is always driven by the same frozen, pretrained locomotion policy which only accepts velocity commands. We use the flat-terrain locomotion policy shipped with Isaac Sim 6.0.1 ({Go2FlatTerrainPolicy}) without modification. We refer the reader to Appendix~\ref{app:sim_params} for the locomotion policy and the physical simulation parameters. All navigation logic above that is written by the LLM (Claude Opus 5 \cite{claudeopus5}). The exact methodology behind how the research literature is utilized and processed differs between our different settings, as described in detail in the following subsections.

\subsection{Literature as context}
\begin{figure}
    \centering
    \includegraphics[width=\linewidth]{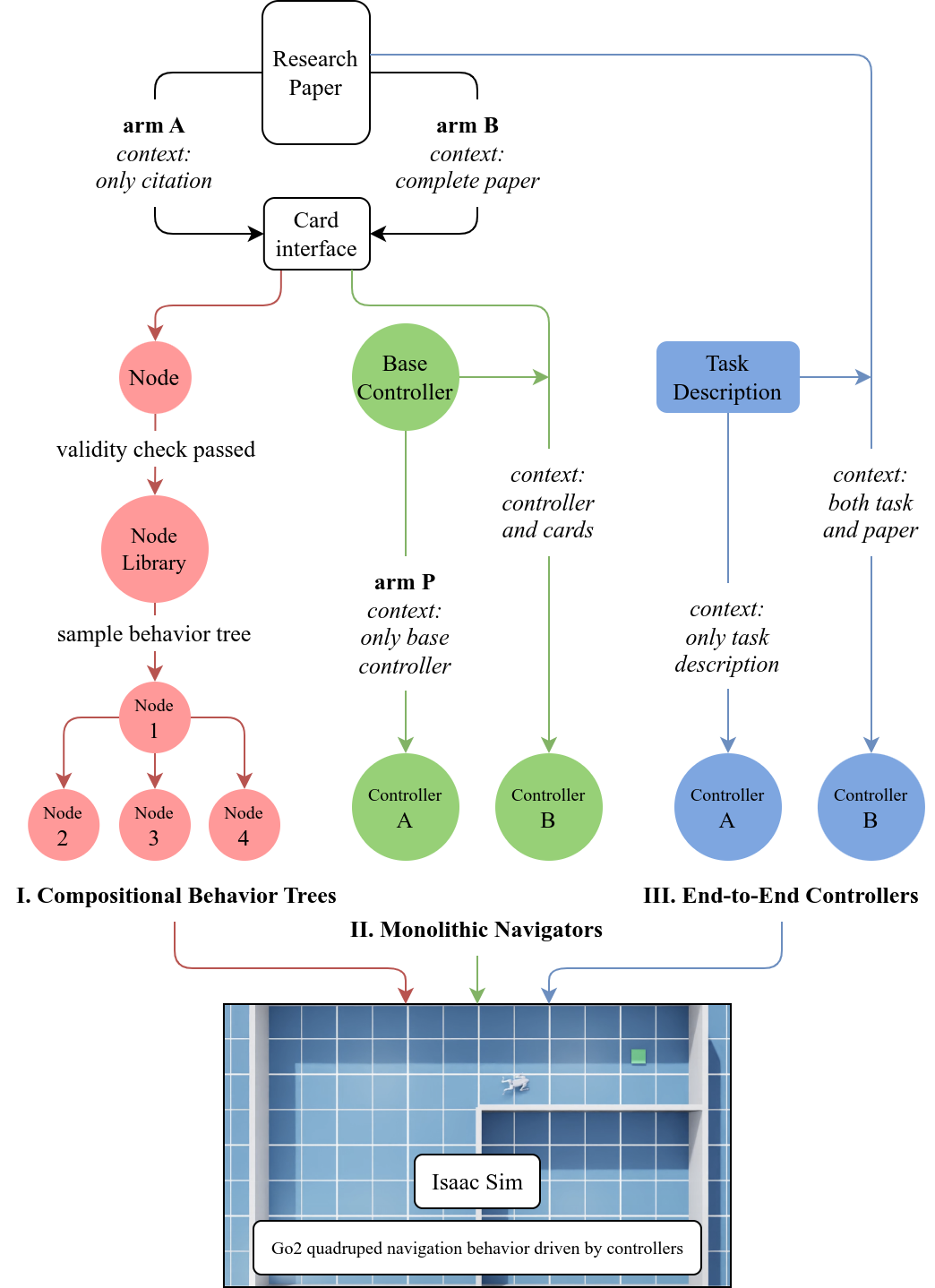}
    \caption{Overview of our three settings. In the first two settings, each hand-picked research paper is condensed into a card, written either from its citation alone (arm $\mathbb{A}$) or from its full text (arm $\mathbb{B}$). \textbf{I. Compositional Behavior Trees:} the LLM writes a behavior node from each card, the nodes that pass a validity check are placed in a library, and behavior trees are sampled from it. \textbf{II. Monolithic Navigators:} the LLM edits a shared base controller using a card. \textbf{III. End-to-End Controllers:} no cards and no base code are used, and the LLM writes a complete controller from the task description alone, or from the task description together with the full text of the paper. In all settings, the resulting controllers drive the high-level navigation of the Go2 robot through velocity commands, with the frozen locomotion policy in Isaac Sim \cite{isaacsim}.}
    \label{fig:pipeline_v1}
\end{figure}

To provide literature context to the LLM, five research papers are carefully selected to cover a wide range of behavioral traits. Specifically, we require each selected paper to describe a mechanism that is separable from its original task and whose trigger could naturally occur in our scenes (e.g., blind corners, pose drift, or tight passages). Furthermore, all five papers are deliberately well known and predate the training cutoff of the model (1986--2004). While novelty therefore cannot be tested in this setup, it ensures that the prior knowledge of the model is high and constant across the papers, isolating what reading a research work adds over merely recalling it. The selected papers cover the following behavioral traits:
\begin{itemize}
    \item Dynamic-window speed limiting \cite{dynamic_window}
    \item Inevitable collision states \cite{fraichard2004inevitable}
    \item Path integration in ants
    \cite{ants} 
    \item Path integration in mammals \cite{mammals}
    \item Geometric reorientation
    \cite{geom_reo}
\end{itemize}
Each of the five papers is condensed into a \textit{card}, serving as an interface between the work it describes and the rest of our pipeline. Cards can also be considered as the abstraction layer above the research work and the primary input to our first two settings. Specifically, every card describes the mechanism of the paper, under which conditions it applies, how it translates to a rule, and what target metrics it aims to improve. We utilize two ways of producing the cards from research works, which we will further refer to as different \textit{arms}. When the LLM is provided only the citation of a paper (authors, title, venue and year) and writes the card from its own memory of the work, we refer to this setting as arm $\mathbb{A}$. Conversely, when it is provided the full text of the paper instead, we refer to it as arm $\mathbb{B}$. The most recent work among our selected papers is dated $2004$, making it very likely that the LLM was exposed to our papers as part of its training data. Such an A/B comparison therefore allows us to test whether reading the full paper leads to better navigation code than writing the card from the model's own memory of it. For the extraction of the cards we utilize Claude Opus 5. For further details on the unified card schema, we refer the reader to Appendix \ref{app:cards}. Finally, the cards from the different arms are processed in the same manner, as shown in Figure \ref{fig:pipeline_v1}.

\subsection{Compositional Behavior Trees}
Initially, our aim is to allow the high-level navigation behavior of the Go2 robot be driven by behavior trees \cite{colledanchise2018bt} composed of nodes. Behavior trees naturally fit our setting, as they decompose a controller into self-contained nodes that can be developed and tested independently, and then combined into larger behaviors without modifying one another. In particular, this allows us to map each research paper to its own node, such that the contribution of every paper remains isolated and attributable. Additionally, since the tree is re-evaluated at every tick, the robot can react to changes in its surroundings by switching between behaviors as their conditions change. Each behavior node encapsulates a target behavior trait, which triggers only when the set of conditions described by the node is met. Altogether, our selected $5$ papers result in a library of $10$ nodes: $5$ from arm $\mathbb{A}$ and $5$ from arm $\mathbb{B}$. The first setting evolves through three stages, each addressing a shortcoming of the previous one.

\subsubsection{Unique behavior tree per arm per task} In the first stage, we produce behavior nodes from a single call. To keep the extracted principle separate from its implementation, we split generation into two calls, first extracting a card and then writing the node from it. In this setting, if the node triggers, it takes full control of the robot. Each arm then builds its unique behavior tree. The results show that arm $\mathbb{A}$ tends to write complete navigators, while arm $\mathbb{B}$ more faithfully implements the specific mechanism the paper describes but often cannot navigate at all. When we remove the paper's mechanism from an $\mathbb{A}$ navigator, it still reaches the goal with a success-weighted path efficiency (SPL) \cite{spl} of $0.93$ and $0.95$ in 2 kinematic rollouts.
\subsubsection{Behavior-tree nodes built from shared base node} Our aim at this stage is to achieve better isolation between the cards and their effects. We achieve this by turning each one into a single node, built on top of a shared base navigator. While this helps with comparability between the two arms, it also splits the results into two extreme groups: $5$ of the $10$ admitted nodes are never active, and $5$ are active on $87.6\%-100\%$ of the ticks. Constantly active nodes effectively override the base navigator and result in worse downstream performance. Specifically, we measure the goal-reaching rate in the kinematic simulator over $20$ generated maze worlds. While the base navigator alone reaches the goal in $10$ of $20$ worlds, none of the $10$ nodes improves on it.   
\subsubsection{Nodes as only extensions to the base navigator} In the third stage, we restrict the lower-level nodes to only extend the base navigator. This restriction, however, does not produce better results. None of the $10$ nodes improves the metric it claims to improve while still completing the task. For example, two nodes that aim to lower the number of collisions only achieve this by barely moving at all, and one of them never even reaches the goal.  Specifically, in the kinematic simulator over the same $20$ maze worlds, where the base navigator alone reaches the goal in $10$ of $20$, none of the $5$ arm $\mathbb{A}$ nodes and none of the $10$ arm $\mathbb{B}$ nodes improve on the base, while $3$ and $4$ of them, respectively, lower it. The only node that helps is the single arm $\mathbb{D}$ node, which raises the goal-reaching rate from $0.50$ to $0.60$, albeit at the cost of more collisions (from $0.40$ to $1.70$).

\subsection{Monolithic Navigators}
In contrast to our behavior tree approach, our second setting tasks the LLM with writing monolithic navigation code to drive the behavior of the Go2 robot. Such a setup, however, requires additional measures to ensure that the comparability between the different arms does not suffer. Specifically, we utilize a shared base navigator, meaning that the arms do not write code from scratch, but instead modify and extend an existing controller. The base navigator itself was written by a neutral model call, receiving neither a card nor a paper, and was frozen and scored before any arm was run. It reaches the goal in $45\%$ of the environment when tested on a randomly generated set of $20$ maze worlds. Its navigation logic was never fixed, to allow room for improvements by the arms. Such a design, however, carries a risk of anchoring. When a model is handed working code, it tends to make conservative edits. A null result from this setting therefore may mean that the cards did not improve this particular navigator, rather than that the research literature does not generally help.
\\
\\
The model is then prompted to fold the cards into the base navigator. However, all $10$ cards describe their mechanism as a behavior-tree node. Therefore, before prompting, the card text is translated out of the behavior-tree vocabulary by a substitution table, without changing any number, and the model is asked to transfer the principle. For this setting, we also add a control arm, $\mathbb{P}$, which receives the same request without any card. While folding the cards into a shared base holds the surrounding code constant across arms, it does not hold the prompt length constant. The prompt of arm $\mathbb{B}$ is about twice as long as that of $\mathbb{P}$ ($2.09\times$). Hence, the difference between $\mathbb{B}$ and $\mathbb{P}$ mixes the content of the cards with the mere amount of added context. The results show that every arm is less competent than the base navigator alone, which reaches the goal in $45\%$ of the $20$ maze worlds. Table~\ref{tab:edit_arms} reports the goal-reaching rate of each of the $k = 3$ samples per arm. $\mathbb{P}$ reaches $20\%$, while $\mathbb{A}$ and $\mathbb{B}$ both reach $16.7\%$, and none of the $9$ controllers outperforms the base. With a standard error of about $0.11$, the differences between the arms are within noise.

\begin{table}[t]
\centering
\caption{Goal-reaching rate of the edited navigators on the $20$ maze worlds. The base navigator alone reaches $0.450$, and the standard error per sample is about $0.11$.}
\label{tab:edit_arms}
\begin{tabular}{lccccc}
\toprule
Arm & $k_0$ & $k_1$ & $k_2$ & Mean & vs.\ base \\
\midrule
$\mathbb{P}$ & $0.050$ & $0.350$ & $0.200$ & $0.200$ & $-0.250$ \\
$\mathbb{A}$ & $0.200$ & $0.050$ & $0.250$ & $0.167$ & $-0.283$ \\
$\mathbb{B}$ & $0.300$ & $0.200$ & $0.000$ & $0.167$ & $-0.283$ \\
\bottomrule
\end{tabular}
\end{table}

\subsection{End-to-End Controllers}
\begin{figure}[t]
    \centering
    \includegraphics[width=\linewidth]{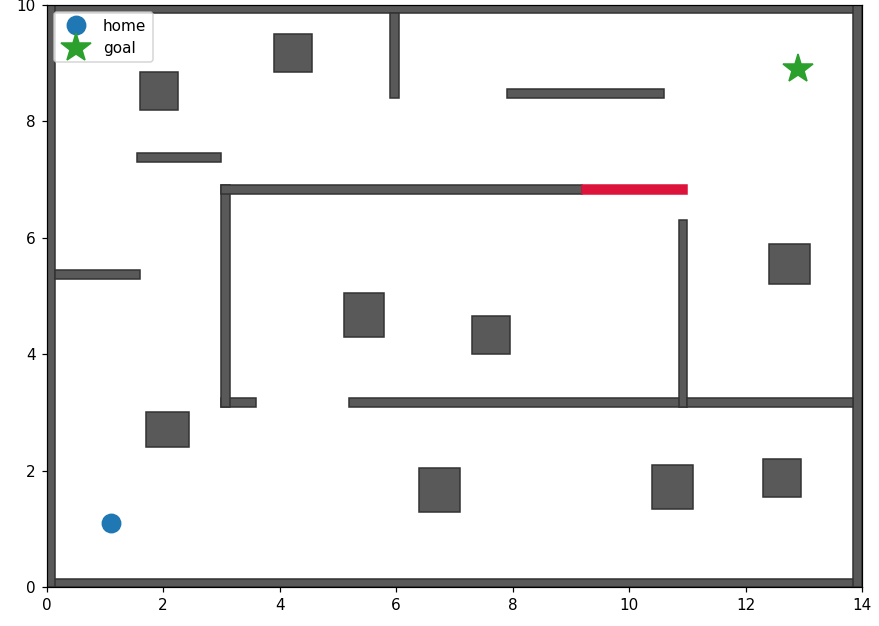}
    \caption{The environment of our third setting. The Go2 robot is tasked to find the goal position (green star), return home (blue circle), and repeat. After the second trip, an obstacle (a door) appears blocking the previously fastest route (highlighted in red).}
    \label{fig:world}
\end{figure}

\begin{figure*}[t]
  \centering
  \includegraphics[width=\textwidth]{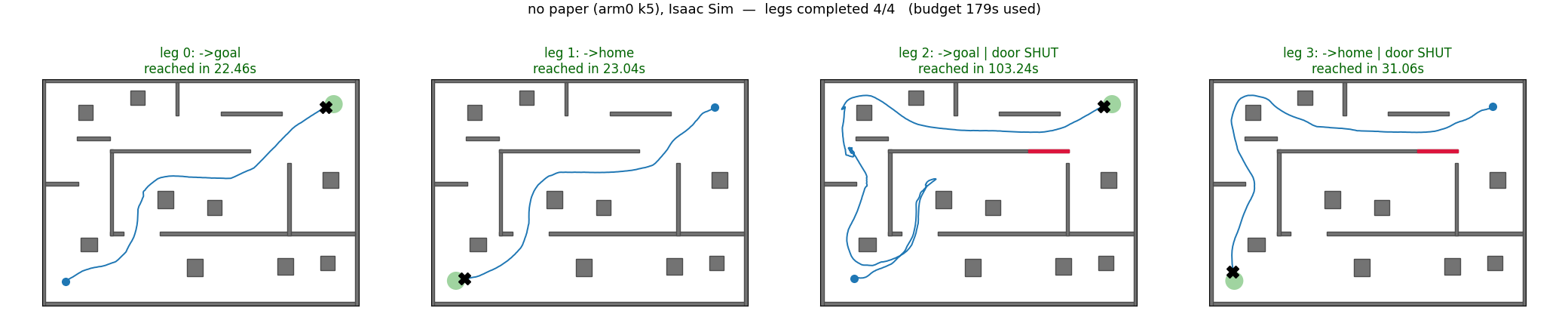}\\[2pt]
  \includegraphics[width=\textwidth]{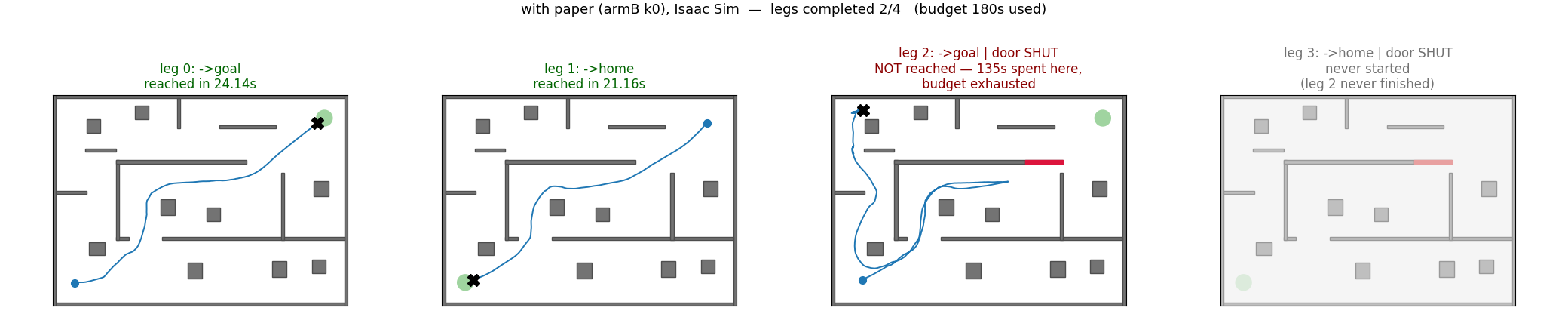}
  \caption{Paths of two controllers in Isaac Sim, one panel per trip.
    Top: without the paper (4/4 trips). Bottom: with the paper (2/4 trips,
    budget exhausted on the third). The obstacle blocking the shortest route from the third trip on is highlighted in red.}
  \label{fig:physics-paths}
\end{figure*}

The third and last setting we designed is aimed at disambiguating the mixed findings of our previous approaches. In this setting, there are no cards and no base code for the model to anchor on. The Go2 robot \cite{unitreego2} is tasked to reach a goal position in a maze-like environment, return home, and repeat. However, the environment changes once during each episode. After the second trip, while the robot is at home, a door on the shortest route shuts, and the robot is forced to find a new route to its goal position. The environment is visualized in Figure \ref{fig:world}, and we refer the reader to Appendix \ref{app:fin_env} for the exact setting and parameters of the environment. As the research paper, we choose D$^*$ (Dynamic A$^*$) \cite{stentz1994optimal}, a well-known incremental-replanning method. It guides the Go2 along the shortest path over a map of everything it has seen so far, and whenever the sensor reveals an obstacle on that path, it updates the map and replans around it. Its main benefit is efficiency, as it repairs the existing plan locally instead of replanning from scratch. While D$^*$ fits our task naturally, it predates the model's training cutoff by decades, and the model can write complete controllers from a task description alone. Rather than introducing new knowledge, the paper therefore tests whether having a finding the model already knows in context changes how it applies that finding to the task. To deal with the run-to-run variance, we generate $k = 8$ controllers per arm. Each controller is the output of a separate generation from the same prompt. The arms are first evaluated in the kinematic simulator, where their performance is indistinguishable ($5/8$ against $6/8$, $p = 1.000$). Then, we evaluate the same $16$ controllers under Isaac Sim physics, where $7/8$ controllers without the paper reference complete all four trips, against $2/8$ with it ($p = 0.041$). A controller counts as successful only if the robot completes all four one-way trips within the allocated $180$\,s time budget. The two prompts used for the controllers are identical apart from the paper reference itself, and neither mentions the challenge of the environment. The exact difference between the two generation setups is given in Appendix \ref{app:prompts}. Table \ref{tab:final} reports how many of the $8$ controllers per arm complete all four trips, while Figure \ref{fig:physics-paths} shows an example controller from each arm under physics. Under physics, controllers written without the paper reference complete on average $3.75$ of the $4$ trips, against $2.38$ for those written with it. In the example pair, after the shortest path is blocked, the controller written without the paper reference finds the detour, while the one written with it stalls and never completes the third trip. However, this does not describe every failure, as one controller written with the paper reference already fails on the second trip, before the path was blocked. Moreover, the two simulators disagree on $6$ of the $16$ controllers: two controllers written without the paper reference fail kinematically but succeed under physics, while four written with it succeed kinematically but fail under physics. Appendix \ref{app:traj} shows an example of the less common outcome of each arm.

\begin{table}[t]
\centering
\caption{Controllers completing all 4 trips, out of 8 independent generations per arm. Arms are compared using a two-sided Fisher's exact test.}
\label{tab:final}
\begin{tabular}{lccc}
\toprule
 & No paper & With paper & Fisher $p$ \\
\midrule
Kinematic simulator & 5/8 & 6/8 & 1.000 \\
Isaac Sim physics & 7/8 & 2/8 & 0.041 \\
\bottomrule
\end{tabular}
\end{table}

\section{Discussion} \label{discussion}
Further experiments and ablations would be required to determine faithfully why the extra literature context does not yield better results. Currently, we maintain two main hypotheses. \\
\newline
\textbf{Hypothesis 1: The model is biased to adopt the research paper's own goals.} In the third setting, the main contribution of the paper is efficiency, as it repairs an existing plan locally, so replanning stays cheap as the map grows. Our trial environment may not be complex enough for this benefit to show, while the paper still introduces complexity in the code and pulls the model toward its own priorities. The cards in our edit-based approach show a similar trend. Each one predicts an improvement in a named metric over a named baseline, and scoring each card for every controller that uses it gives $60$ predictions. Against the baseline named by the card itself, $43$ of $60$ ($72\%$) come true. However, against the base navigator the controller actually replaces, only $10$ of $60$ come true, and in each of these cases the controller reaches the goal less often. In other words, not a single card improves its metric without hurting the task.\\
\newline
\textbf{Hypothesis 2: Code shaped by the paper is more fragile under simulated physics.} Three of the eight controllers written with the paper implement D$^*$ only incrementally, with no fallback to a full replan, and all three fail under physics, although two other failing controllers do include a full replan. One possibility is that a locally repaired search tree can go stale and keep the robot on a route that no longer exists, whereas code that recomputes everything each step cannot hold on to an outdated belief. Another possibility is that the controllers written without the paper choose velocities by simulating short rollouts, which copes better when the gait does not go exactly where it was told, while following the backpointers of D$^*$, which link each map cell to the next one on the path to the goal, does not.

\section{Conclusion}
To conclude, in our tested settings, adding research-paper context does not improve navigation performance, as none of the $10$ nodes and none of the edited navigators built from the cards outperform the shared base navigator. In our third setting, the arms are indistinguishable in the kinematic simulator. Under physics, however, controllers written with the paper complete all four trips less often than those written without it ($2/8$ against $7/8$, $p = 0.041$). Therefore, we formulate our answer to the research question as follows: \\ \newline
\textit{Within the tasks, papers and model we tested, providing research-paper context does not increase the task-completion rate of the LLM-written navigation code for the simulated Go2 robot, and in some cases lowers it.}

\section{Limitations and Threats to Validity}
\noindent\textbf{Limitations.} The design of the evaluation pipeline is inherently limited by comparability. In particular, maintaining comparability to our baselines while keeping the scope of this work relevant for potential real-world adoption remains a challenge throughout the work. In current agentic development settings, the model works over a long, changing set of instructions, often in collaboration with other agents or with a human in the loop. Such a setup, however, would no longer be comparable to the baseline due to high output variance between the different runs.\\
\newline
\noindent\textbf{Threats to Validity.}
Without a settable temperature parameter, the model outputs are non-deterministic. While keeping the model configuration and prompt fixed, we observe an output variance of $62\%$ ($354$ lines) in the length of the generated controllers. Our primary approach to combat variance is to aggregate results over multiple generated controllers per arm, making the high variance statistically less relevant. Furthermore, from a design perspective, several of our settings build the controllers around a shared base. This helps with comparability as the agent is guided to modify the existing shared code only when the new implementation really demands it. In the case of behavior trees, the agent is even unable to modify this base navigator, as every other node was built directly on top of it.
\section{Future Work}
Our hypotheses discussed in Section \ref{discussion} point to three future directions. To begin with, future work could explore running the experiments in more complex environments (e.g., larger rooms with more obstacles and long-horizon objectives), where the efficiency gains from the research papers could be more relevant. Furthermore, adding a fallback to a full replan would test directly whether the fragile behavior of arm $\mathbb{B}$ comes from the locally repaired plans. Finally, constructing the library from research papers published after the release of the model in use could highlight potential benefits of the pipeline. 

\section*{Acknowledgments}
This work was supported by the Mindfire Foundation. We thank them for their funding and for the feedback they gave throughout the project.

\section*{Generative AI Declaration}
The authors acknowledge the use of Anthropic Claude and autonomous development agents to assist with code scaffolding for simulation and evaluation scripts, automated literature cross-referencing, and grammatical refinement of the manuscript. All AI-assisted contributions and bibliographic entries were reviewed and verified for correctness by the authors. All analyses, experimental designs, and final conclusions were authored and verified by the authors.

\bibliography{main}
\bibliographystyle{IEEEtran}

\clearpage
\appendices
\onecolumn

\section{Prompt Templates and Context Injections}
\label{app:prompts}

Every controller of the third setting is the output of a single call to Claude Opus 5 (model ID \texttt{claude-opus-5}) through the Claude Code CLI, version 2.1.278. This appendix shows how the two arms of the third setting differ, namely the block through which the research paper is injected into the prompt. 
Both arms receive the same task description, and the with-paper arm additionally receives the block below, where \texttt{\{\{TASK\}\}} marks the task description and \texttt{\{\{PAPER\}\}} the paper as plain text. Removing the entire block gives the no-paper prompt, so the two prompts are otherwise byte-identical.

{\footnotesize
\begin{verbatim}
{{TASK}}

<!-- ---- PAPER BLOCK BEGIN ---- -->

## Source material

The full text of a published paper is reproduced below. **Use what it
establishes.** Read it, work out which of its ideas bear on the task above, and
build them into your controller where they help.

Work from the document, not from your memory of it. Where the two disagree, the
document wins. The paper was written about something other than a Go2 with a
planar lidar, so carrying it across is your job -- keep what transfers, and do
not force what does not.

Add a short comment at the top of your file naming what you took from the paper
and where in it that came from.


```
{{PAPER}}
```

<!-- ---- PAPER BLOCK END ---- -->
\end{verbatim}
}

\section{Card Template and Example Cards}
\label{app:cards}

Each arm uses the same card schema, which has six fields, as defined in Table~\ref{tab:card-schema}. Additionally, each card also carries an \textit{id} and a \textit{node\_name}. The prompt that turns a card into code receives no indication of which arm produced it.

\begin{table}[!ht]
\centering
\caption{The unified principle-card schema.}
\label{tab:card-schema}
\small
\begin{tabular}{@{}lp{0.70\textwidth}@{}}
\toprule
Field & Content \\
\midrule
source & citation, and a span: the section,
page, figure or equation where the mechanism is stated, left empty when
the writer cannot point at one. A span written by a model is a claim and
is marked unverified. \\
mechanism & What happens and why it helps: the finding, not the
implementation. \\
trigger & The condition under which it applies, which becomes the
precondition. \\
derivation & How the finding becomes a rule for this robot and
these sensors, with numbers, units and the reasoning behind them. \\
prediction & statement (a falsifiable claim),
metric (from a closed list the scorer computes),
baseline, min\_effect (smallest effect that counts as
support) and cost\_metric (the metric expected to get worse). \\
confidence & high / medium / low,
defined by what the source states \\
\bottomrule
\end{tabular}
\end{table}

\clearpage
\section{Third Setting Environment Configuration}
\label{app:fin_env}

Table~\ref{tab:fin-env} lists the task, the world and the controller
interface. All lengths are computed with the benchmark's own geometry
code, inflating obstacles by the 0.30\,m robot radius. The same 2D world
description is used in both simulators.

\begin{table}[!ht]
\centering
\caption{Third setting task, world and controller interface.}
\label{tab:fin-env}
\small
\begin{tabular}{@{}lp{0.62\textwidth}@{}}
\toprule
Setting & Value \\
\midrule
World & Hand-authored building, $14 \times 10$\,m, walls 0.15\,m thick;
no map is provided to the controller \\
Home, goal & $(1.10, 1.10)$ and $(12.90, 8.90)$\,m \\
Task & Four one-way trips: home $\to$ goal, goal $\to$ home,
home $\to$ goal, goal $\to$ home. The harness sets the current target and
switches it on arrival. \texttt{reset()} is called once, before trip 1 \\
Door event & One door ($1.80 \times 0.15$\,m) on the short route is open
for trips 1--2 and shuts once, after trip 2, while the robot is at home \\
Shortest route & 15.03\,m with the door open; 19.36\,m with it shut
(+4.33\,m). The narrowest point of the detour leaves the robot's centre
0.77\,m of free width \\
Time budget & 180\,s of simulated time for all four trips together \\
Arrival radius & 0.40\,m around the current target \\
Success & All four trips completed within the budget \\
Control rate & 50\,Hz (\texttt{step()} called every 0.02\,s) \\
Lidar & 181 beams over $360^\circ$, maximum range 6.0\,m ($\infty$
where nothing returns), computed analytically by ray-casting against the
2D world in both simulators \\
Pose & Exact pose estimate $(x, y, \psi)$, no drift.
pose\_error\_estimate is 0 \\
Other observations & Episode time, a contact flag, and the current
target point \\
Command & Body-frame $(v_x, v_y, \omega_z)$, saturated to
$|v_x|, |v_y| \le 0.8$\,m/s and $|\omega_z| \le 1.0$\,rad/s \\
Imports allowed & Python standard library (minus process, file-system and
network modules), NumPy, and the command type. Anything else is rejected
before the controller runs \\
\midrule
Kinematic simulator & Body-frame velocity integrated exactly at 50\,Hz;
the robot is a disc of radius 0.30\,m. A step is refused only if it would
bring the centre within 0.165\,m of an obstacle. Start pose home, heading
0.6\,rad. Three repetitions per controller \\
Physics simulator & Isaac Sim 6.0.1 with the frozen
\textit{Go2FlatTerrainPolicy} in the loop (Appendix~\ref{app:sim_params});
the body is $0.70 \times 0.31$\,m. One episode per controller in the
pilot \\
\bottomrule
\end{tabular}
\end{table}

\section{Physical Simulation Parameters}
\label{app:sim_params}

Table~\ref{tab:sim-params} lists what our code sets or records for the
physics runs. Everything not listed is left at the defaults of the Isaac
Sim Go2 policy example and robot asset.

\begin{table}[!ht]
\centering
\caption{Physics-simulation settings set or recorded by our code.}
\label{tab:sim-params}
\small
\begin{tabular}{@{}lp{0.62\textwidth}@{}}
\toprule
Setting & Value \\
\midrule
Simulator & NVIDIA Isaac Sim 6.0.1, Python 3.12 \\
Locomotion policy & \textit{Go2FlatTerrainPolicy}, the pretrained
flat-terrain policy shipped with Isaac Sim. module
\textit{isaacsim.robot.policy.examples.robots.go2} \\
Robot asset & Isaac/Robots/Unitree/Go2/go2.usd from the Isaac
Sim asset library \\
Physics step & 0.005\,s (200\,Hz) \\
Decimation & 4: each velocity command is held for 4 physics steps \\
Spawn and settling & Spawned at home, 0.55\,m above the floor, heading
0.6\,rad. Also, the robot receives 2.0\,s of zero command to stand before the episode clock
starts \\
Scene & Exported from the 2D floor plan \\
Door & A single prim, moved below the floor for trips 1--2 and back
to its place for trips 3--4 \\
Sensing & Lidar ray-cast analytically against the 2D world. The pose is
read from the simulated robot base. No rendered sensor is used \\
Rendering & Off for the scored runs; rendering never advances physics \\
Hardware & Local workstation, NVIDIA RTX 4060 Laptop GPU (8\,GB)\\
Wall clock & 170--305\,s per 180\,s episode \\
\bottomrule
\end{tabular}
\end{table}

\section{Per-Controller Results}
\label{app:controllers}

Table \ref{tab:per-controller} lists every controller of our third setting. The six controllers whose verdict differs between the two simulators are no-paper k2 and k5, which succeed only under physics, and with-paper k1, k2, k6 and k7, which succeed only kinematically.

\begin{table}[!ht]
\centering
\caption{Trips completed by each controller of our third setting, in the three kinematic repetitions and the single physics episode of Table \ref{tab:final}. Lines: length of the generated file. Bold: all four trips, which counts as a success. The shared base navigator of our earlier settings is given for reference.}
\label{tab:per-controller}
\small
\begin{tabular}{@{}llrcc@{}}
\toprule
Arm & $k$ & Lines & Kinematic (3 reps) & Physics \\
\midrule
No paper & k0 & 658 & \textbf{4}, \textbf{4}, \textbf{4} & \textbf{4} \\
 & k1 & 770 & \textbf{4}, \textbf{4}, \textbf{4} & \textbf{4} \\
 & k2 & 815 & 3, 3, 3 & \textbf{4} \\
 & k3 & 651 & \textbf{4}, \textbf{4}, \textbf{4} & \textbf{4} \\
 & k4 & 760 & \textbf{4}, \textbf{4}, \textbf{4} & \textbf{4} \\
 & k5 & 740 & 2, 2, 2 & \textbf{4} \\
 & k6 & 851 & \textbf{4}, \textbf{4}, \textbf{4} & \textbf{4} \\
 & k7 & 758 & 3, 3, 3 & 2 \\
\cmidrule(l){2-5}
 & \multicolumn{2}{l}{Successes} & 5/8 & 7/8 \\
\midrule
With paper & k0 & 869 & 2, 2, 2 & 2 \\
 & k1 & 781 & \textbf{4}, \textbf{4}, \textbf{4} & 2 \\
 & k2 & 922 & \textbf{4}, \textbf{4}, \textbf{4} & 2 \\
 & k3 & 659 & \textbf{4}, \textbf{4}, \textbf{4} & \textbf{4} \\
 & k4 & 568 & \textbf{4}, \textbf{4}, \textbf{4} & \textbf{4} \\
 & k5 & 599 & 2, 1, 2 & 2 \\
 & k6 & 636 & \textbf{4}, \textbf{4}, \textbf{4} & 1 \\
 & k7 & 888 & \textbf{4}, \textbf{4}, \textbf{4} & 2 \\
\cmidrule(l){2-5}
 & \multicolumn{2}{l}{Successes} & 6/8 & 2/8 \\
\midrule
Shared base & -- & 577 & 2, 2, 2 & 3 \\
\bottomrule
\end{tabular}
\end{table}

\clearpage
\section{Trajectories of the Less Common Outcomes}
\label{app:traj}
\begin{figure}[h]
\centering
\includegraphics[width=\columnwidth]{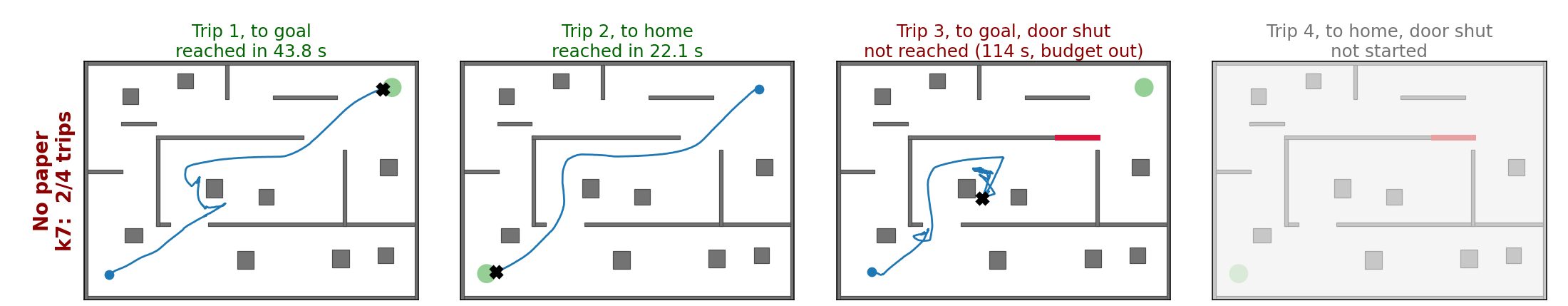}\\[2pt]
{\small (a) Written without the paper: 2 of 4 trips completed.}\\[8pt]
\includegraphics[width=\columnwidth]{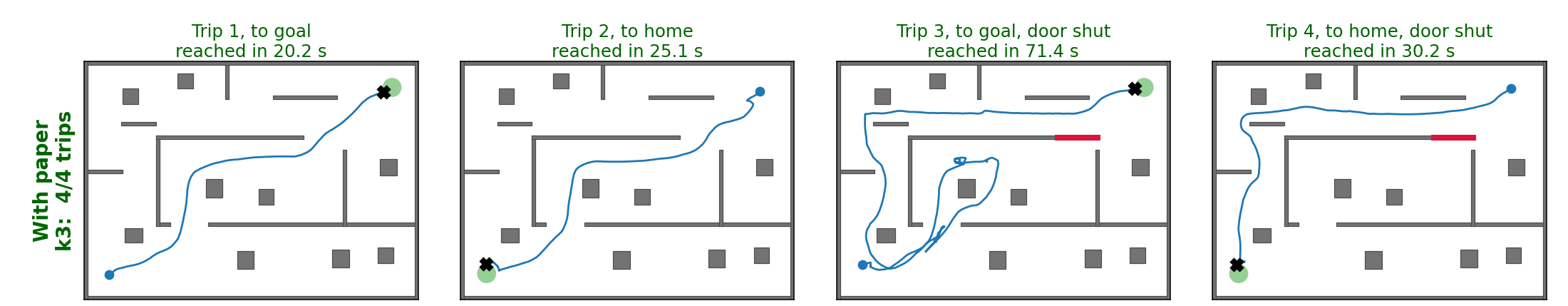}\\[2pt]
{\small (b) Written with the paper: 4 of 4 trips completed.}
\caption{Trajectories under Isaac Sim physics of the less common outcome of each arm.}
\label{fig:traj-exceptions}
\end{figure}

Figure \ref{fig:physics-paths} shows the most common outcome of each arm under physics. Figure \ref{fig:traj-exceptions} complements it with the less common one. The controller in (a) is the only one of the 8 written without the paper that fails under physics. It completes the first two trips, but it is unable to complete the third trip after the door shut. In the kinematic simulator, the same controller completes three of the four trips. The controller in (b) is one of the two written with the paper that succeed under physics, completing all four trips, as it also does in the kinematic simulator.

\end{document}